\documentclass[a4paper,fleqn]{cas-sc}

\usepackage[square, comma, sort&compress, numbers]{natbib}

\usepackage{amsmath}
\usepackage[linesnumbered,boxed]{algorithm2e}
\usepackage[justification=centering]{caption}
\usepackage{caption}
\usepackage{subcaption}
\usepackage{graphicx}
\usepackage{longtable}
\usepackage{multirow}
\usepackage{array}
\usepackage{pdflscape}
\usepackage{booktabs}
\usepackage[figuresright]{rotating}
\usepackage{multirow}

\newproof{pf}{Proof}
\usepackage{lineno}
\usepackage{float}
\def\tsc#1{\csdef{#1}{\textsc{\lowercase{#1}}\xspace}}
\tsc{WGM}
\tsc{QE}
\tsc{EP}
\tsc{PMS}
\tsc{BEC}
\tsc{DE}

\begin{document}
\let\printorcid\relax
\let\WriteBookmarks\relax
\def\floatpagepagefraction{1}
\def\textpagefraction{.001}

\title [mode = title]{ Toward Dependency Dynamics in Multi-Agent Reinforcement Learning for Traffic Signal Control}

\author[1]{Yuli Zhang}[]
\ead{Yuli.Zhang20@student.xjtlu.edu.cn}


\author[2]{Shangbo Wang\textsuperscript{*}}  
\ead{Shangbo.Wang@sussex.ac.uk}  
\author[1]{Dongyao Jia\textsuperscript{*}}  
\ead{Dongyao.Jia@xjtlu.edu.cn} 
\author[1]{Pengfei Fan}  
\ead{Pengfei.Fan22@student.xjtlu.edu.cn} 
\author[1]{Ruiyuan Jiang}  
\ead{Ruiyuan.Jiang20@student.xjtlu.edu.cn} 
\author[1]{Hankang Gu}  
\ead{Hankang.Gu16@student.xjtlu.edu.cn} 
\author[3]{Andy H.F. Chow}  
\ead{andychow@cityu.edu.hk} 

\cortext[cor1]{Corresponding authors: Shangbo Wang\textsuperscript{b}\textsuperscript{*} and Dongyao.Jia\textsuperscript{a}\textsuperscript{*}}

\affiliation[1]{organization={School of Advanced Technology},
	addressline={Xi'an Jiaotong-Liverpool University}, 
	city={Suzhou},
	postcode={215123}, 
	country={China}}

\affiliation[2]{organization={Department of Systems Engineering},
	addressline={University of Sussex}, 
	city={Brighton},
	postcode={BN1 9RH},
	country={UK}}

\affiliation[3]{organization={Department of Engineering and Design},
	addressline={City University of Hong Kong}, 
	city={Hong Kong},
	postcode={SAR},
	country={China}}

\newcommand{\figref}[1]{Fig.~\ref{#1}}

\begin{abstract}
Abstract--- Reinforcement learning (RL) emerges as a promising data-driven approach for adaptive traffic signal control (ATSC) in complex urban traffic networks, with deep neural networks substantially augmenting its learning capabilities. However, centralized RL becomes impractical for ATSC involving multiple agents due to the exceedingly high dimensionality of the joint action space. Multi-agent RL (MARL) mitigates this scalability issue by decentralizing control to local RL agents. Nevertheless, this decentralized method introduces new challenges: the environment becomes partially observable from the perspective of each local agent due to constrained inter-agent communication. Both centralized RL and MARL exhibit distinct strengths and weaknesses, particularly under heavy intersectional traffic conditions. In this paper, we justify that MARL can achieve the optimal global Q-value by separating into multiple IRL (Independent Reinforcement Learning) processes when no spill-back congestion occurs (no agent dependency) among agents (intersections). In the presence of spill-back congestion (with agent dependency), the maximum global Q-value can be achieved by using centralized RL. Building upon the conclusions, we propose a novel Dynamic Parameter Update Strategy for Deep Q-Network (DQN-DPUS), which updates the weights and bias based on the dependency dynamics among agents, i.e. updating only the diagonal sub-matrices for the scenario without spill-back congestion. We validate the DQN-DPUS in a simple network with two intersections under varying traffic, and show that the proposed strategy can speed up the convergence rate without sacrificing optimal exploration. The results corroborate our theoretical findings, demonstrating the efficacy of DQN-DPUS in optimizing traffic signal control. We applied the proposed method to a dual-intersection, and the results indicate that our approach performs effectively under various traffic conditions. These findings confirm the robustness and adaptability of DQN-DPUS in diverse traffic densities, ensuring improved traffic flow and reduced congestion.
\end{abstract}

\begin{highlights}
	\item Analyze spill-back effects impacting MARL dependency dynamics at intersections.
	\item Prove optimal Q-values can be achieved without scalability challenges in no-spill-back cases.
	\item Propose DQN-DPUS, blending centralized and distributed learning for robust MARL.
	\item Validate DQN-DPUS on SUMO, showing superior performance with increased congestion.
\end{highlights}





\maketitle

\section{Introduction}
\noindent Traffic congestion has emerged as a perplexing issue in urban areas, primarily stemming from the challenge of efficiently utilizing the limited road infrastructure. Traffic signal control can effectively mitigate congestion by optimizing the traffic signal timing parameters at signalized intersections without major changes to the existing infrastructure. Current deployed traffic light control systems often rely on fixed schedules that fail to account for real-time traffic conditions or only do so minimally. In contrast, adaptive traffic signal control (ATSC) at intersections is crucial in optimizing road resource utilization and alleviating traffic congestion by dynamically regulating traffic flow \cite{1}. Many classical ATSC systems have been created and extensively implemented worldwide. Existing Traffic Signal Control (TSC) methods can be broadly classified into classical Adaptive ATSC systems and RL-based ATSC systems. Classical ATSC systems, such as the Split Cycle Offset Optimization Technique (SCOOT) \cite{2}, Chiu's distributed adaptive control system \cite{3}, Zhang's fuzzy logic controller (FLC) \cite{4}, Webster's method \cite{5}\cite{6}, and MaxBand \cite{7}, compute optimal signal plans based on various traffic parameters like demand and saturation rate. While classical ATSC systems are widely used due to their responsiveness to real-time traffic, they are often complex, nonlinear, and stochastic, which makes it challenging to find optimal signal settings. To address these computational challenges, reinforcement learning (RL) techniques have been widely applied to ATSC. For instance, Prashanth and Bhatnagar (2009) \cite{8} introduced RL with function approximation for ATSC, which reduced computational overhead. Similarly, Cai et al. (2009) applied approximate dynamic programming (ADP) and RL to further enhance signal control \cite{9}. RL-based approaches for ATSC have shown significant advantages, such as high computational efficiency, the ability to adapt to changing traffic patterns in real time, and scalability to various traffic conditions \cite{new3}. However, despite these improvements, RL-based ATSC still faces limitations, such as relying solely on vehicles in the current phase, which constrains the optimal use of resources across entire intersections \cite{10}.

Recently, deep reinforcement learning (DRL) has made significant advancements by leveraging deep neural networks. Techniques such as Deep Q-Networks (DQN) \cite{11}, Advantage Actor-Critic (A2C) \cite{12}, and Deep Deterministic Policy Gradient (DDPG) \cite{13} have improved RL’s ability to approximate value functions for ATSC. Furthermore, multi-agent RL (MARL) has demonstrated considerable potential in managing complex environments by improving coordination and scalability in large-scale traffic networks, allowing for more effective decision-making through extensive training and real-time feedback.

In MARL, the state observations and rewards for multiple intersections are interdependent, as they are influenced by the actions taken by other intersections, and all intersections share a common optimization goal; Existing MARL technique can be generally classified into three categories: fully centralized learning \cite{14}\cite{15}, fully decentralized learning and hybrid centralized learning and decentralized execution \cite{16}\cite{18}. One of the key challenges encountered by most existing MARL algorithms is their tendency to introduce issues such as non-stationarity, scalability limitations, and communication overhead. These challenges stem from the concurrent learning of multiple agents, the increasing number of agents, and the communication demands within MARL systems \cite{19}.  

To alleviate the aforementioned challenges of MARL, some researchers have applied techniques such as centralized training with decentralized execution (CTDE) and attention-based communication mechanisms to address non-stationarity and scalability issues \cite{20}\cite{21}. Non-stationarity in MARL is primarily caused by the concurrent learning of agents, which indirectly leads to inter-agent dependencies. In multi-agent systems, such dependencies can vary over time, creating dynamic dependency. For example, in cooperative navigation tasks, the behavior of one agent can dynamically influence the optimal strategies of others. This raises an important question: how can such dynamic dependencies be leveraged to mitigate non-stationarity in MARL?

%
%

\begin{figure}[t]
	\centering
	\includegraphics[width=0.7\linewidth]{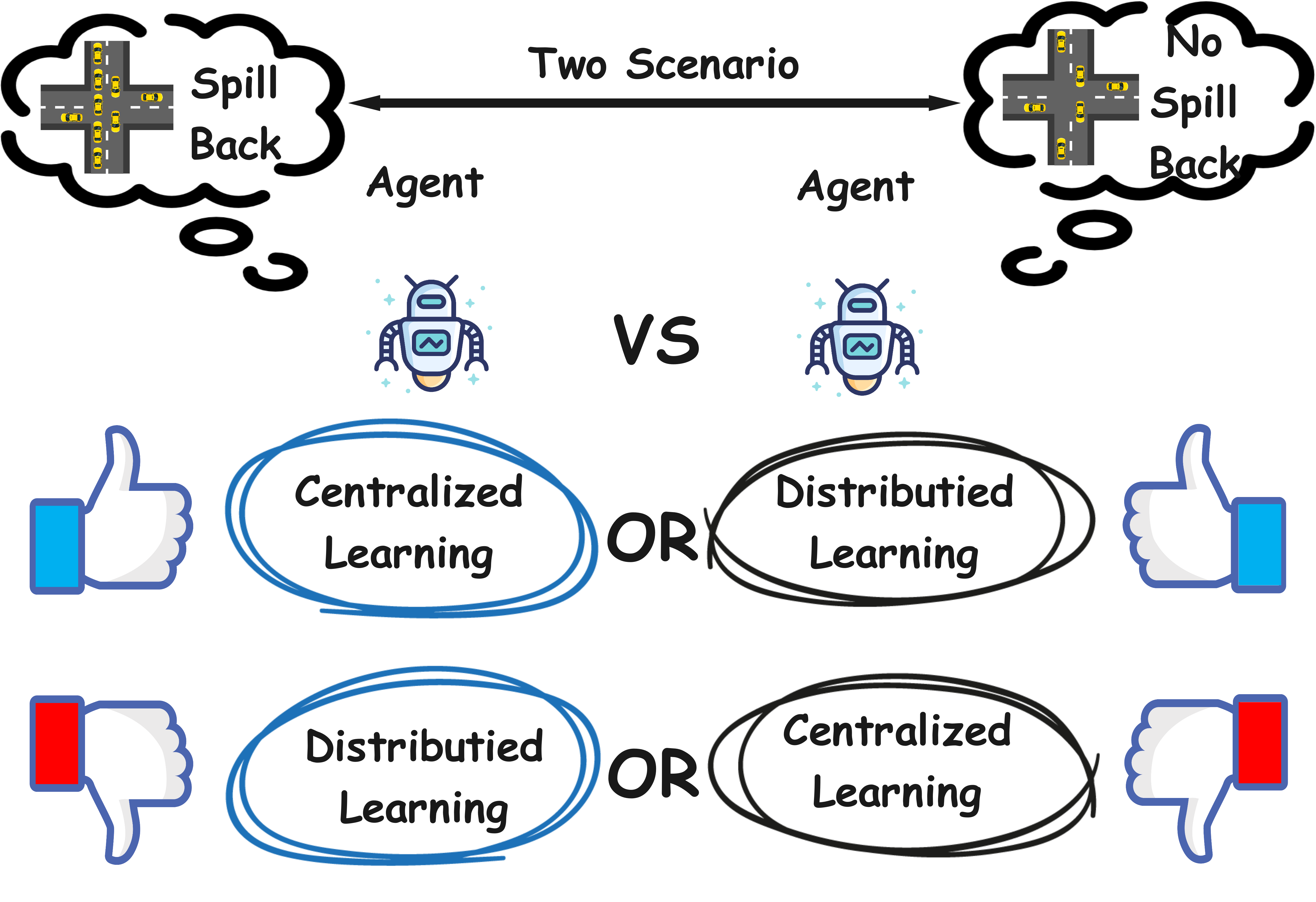}
	\caption{Illustration of two scenarios for traffic management: with and without spill-back. In the spill-back scenario, centralized learning is preferred, while distributed learning is more effective in the no-spill-back scenario. Centralized learning coordinates actions among agents to manage congestion, whereas distributed learning allows agents to act independently, improving efficiency without congestion. The choice of algorithm is crucial in different traffic conditions.}
	\label{fig1}
\end{figure}

In this paper, we aim to address the proposed question by investigating how to leverage dynamic dependencies in multi-agent systems to improve traffic signal control (TSC). In cases of low traffic density, the optimal approach involves decentralized, independent control at each intersection. Conversely, in regions with high traffic density, cooperation among signals in that specific area becomes imperative for optimizing overall traffic flow \cite{new10}. Fig. \ref{fig1} illustrates the challenge of achieving a balance that allows the entire system to benefit from both the scalability of MARL and the optimality of fully centralized RL techniques simultaneously. In this paper, we aim to investigate how to utilize the dynamics of dependency (spill-back) among agents (intersections) to mitigate the challenges of the non-scalability, non-stationarity and communication overhead in MARL based TSC. We justify that MARL can achieve the optimal global Q-value by separating into multiple IRL (Independent Reinforcement Learning) processes when no spill-back effect occurs (no dependency) among intersections (agents). Based on this conclusion, we propose a novel \textbf{Dynamic Parameter Update Strategy} (DPUS), which dynamically updates the network parameters based on the dependency dynamics among agents. More specifically, the main contributions of this paper are listed as follows: 

1) We show that the dependency dynamics in MARL based TSC is caused by the time-varying spill-back effect occurred between adjacent intersections, and provide an analysis about the impact of spill-back effect on the MARL learning process.

2) Through theoretical proofs, we demonstrate that in the absence of spill-back during whole time horizon, IRL can achieve the same performance as fully centralized learning. This finding gives the sufficient condition for IRL achieving the optimal Q-value without introducing the non-stationarity, scalability challenges.

3) We propose a dynamic parameter update strategy for Deep Q-Network (DQN-DPUS), which updates the weights and bias based on the dependency dynamics among agents, i.e. updating only the diagonal sub-matrices for the scenario without spill-back effect. Additionally, DQN-DPUS combines the advantages of centralized learning methods and fully distributed algorithms, leveraging the strengths of both approaches to achieve efficient and robust learning in multi-agent environments.

4) We validate the DQN-DPUS algorithm on the SUMO simulation platform for the scenario with two traffic intersections and varying input vehicle densities. The experimental results demonstrate that as the level of congestion in the road network increases, DQN-DPUS outperforms the baselines. 

The rest of the paper is arranged as follows: Section \ref{S2} discusses the related work. Section \ref{S3} introduces the MDP and MAMDP. Section \ref{S4} introduces dependency dynamics in MARL-based TSC. Section \ref{S5} explains our DPUS-DQN algorithm. Section \ref{S6} describes the experiment settings. Section \ref{S7} discusses the results and section \ref{S8} concludes this paper. 

\section{Related Works} \label{S2}
In this section, we conduct a literature review of existing MARL based ATSC, which are generally classified into centralized RL and decentralized RL.
\subsection{Centralized RL}
Centralized RL methods involved evaluating the returns of actions for all agents to derive the optimal joint strategy across all traffic intersections. However, in large-scale traffic networks, these methods suffered from the curse of dimensionality due to the exponential growth of the action space as the number of intersections increased. To address this issue, Van der Pol and Oliehoek \cite{22} proposed decomposing the global Q-function into a linear combination of local subproblems. To ensure that individual agents considered the learning processes of other agents, Tan et al. \cite{23} modeled the joint Q-function as a weighted sum of local Q-functions, minimizing deviations from the global return. Furthermore, Zhu et al. \cite{24} introduced a joint tree within a probabilistic graph model to facilitate the computation of joint action probabilistic reasoning. Joint action learning was a critical area in multi-agent reinforcement learning (MARL), particularly in methods based on centralized training and decentralized execution \cite{25}\cite{26}. However, the scalability issue became evident with the growth of the number of agents involved \cite{new7}.

Efforts to scale up the number of agents included the approach by Wang et al. \cite{27}, who introduced a cooperative double Q-learning method for the ATSC problem. The method utilized mean-field approximation \cite{28} to model interactions among agents, where the joint actions of other agents were averaged into a scalar using mean-field theory, thereby reducing the dimensionality of the agents' action space in large-scale environments. Compared to centralized RL methods, decentralized RL was more widely used in ATSC.

\subsection{Decentralized RL}
In the field of decentralized RL learning for traffic signal control, each agent autonomously managed a specific intersection, typically with only partial observation of the entire environment. Collaboration among these agents mainly occurred through the exchange of observations and policies \cite{29}. Research efforts were directed towards developing methods that derived comprehensive global state features from local information exchanges and interactions among intersections. For example, the MA2C algorithm \cite{30} extended the independent A2C algorithm to multi-agent scenarios by incorporating state information and strategies from neighboring agents. Similarly, Wei et al. \cite{31} integrated the max-pressure approach \cite{32} into multi-agent RL to achieve a more intuitive representation of state and reward functions.

Despite these advancements, decentralized RL methods often faced challenges related to coordination and scalability, especially in complex traffic scenarios where optimal strategies for intersections could vary significantly. To address these issues, Chen et al. \cite{33} proposed specifying individual rewards for each intersection to capture the coordination demands between neighboring intersections. Zhang et al. \cite{34} introduced a neighborhood cooperative Markov game framework, defining the goal of each intersection as the average accumulated return within its neighborhood and independently learning cooperative strategies based on the 'lenient' principle. Wang et al. \cite{35} presented a decentralized framework based on A2C, where global control was assigned to each local RL agent. In this setup, global information was constructed by concatenating observations (state and reward information) from neighboring intersections, allowing agents to consider local interactions while making decisions.

Ma and Wu \cite{36} extended MA2C with a hierarchical approach by dividing the traffic network into regions, each managed by a high-level agent, while low-level agents controlled the traffic lights within those regions. However, these methods had limitations in dynamic traffic environments. The varying distribution of vehicles at different intersections affected the performance of these approaches, as they did not adapt quickly to changing traffic conditions \cite{new8}. Additionally, mechanisms such as attention models used to estimate the correlation between intersections may not have been sufficiently sensitive to fluctuations in traffic patterns \cite{new9}.

Existing approaches often struggled to effectively balance coordination and scalability in dynamic traffic networks. The main limitations are insufficient adaptability to dynamic traffic distributions and the overhead associated with communication and computation when coordinating multiple agents. These challenges led to suboptimal performance, especially under rapidly changing traffic conditions. 

\section{PRELIMINARIES}\label{S3}
To enhance the readability, we firstly review the fundamentals of single-agent and multi-agent RL.
\subsection{Single-Agent Markov Decision Process}
A Markov Decision Process (MDP) is the foundational framework for illustrating reinforcement learning (RL) in a single-agent setting. It is formally defined as a 5-tuple $\langle \mathcal{S}, \mathcal{A}, \mathcal{P}, \mathcal{R}, \gamma \rangle$. Here, $\mathcal{S}$ represents the state space, and $\mathcal{A}$ denotes the action space. The transition function $\mathcal{P}: \mathcal{S} \times \mathcal{A} \times \mathcal{S} \rightarrow [0, 1]$ maps a state-action pair to a new state, reflecting the environment's dynamics. The reward function $\mathcal{R}: \mathcal{S} \times \mathcal{A} \rightarrow \mathbb{R}$ provides feedback to the agent based on its actions. The discount factor $\gamma \in [0, 1]$ discounts future rewards to account for the uncertainty in long-term predictions.

Formally, the objective of reinforcement learning in an MDP is to find a policy $\pi: \mathcal{S} \rightarrow \Delta(\mathcal{A})$ that maximizes the expected discounted return. This is quantified by the state-value function:
\begin{equation}\label{eq1}
	V^\pi(s) = \mathbb{E}^\pi \left[ \sum_{k=0}^{\infty} \gamma^k \mathcal{R} \left( s^{(t)}, \pi \left( s^{(t)} \right) \mid s_0 = s \right) \right]
\end{equation}
where $\pi$ maps states to a probability distribution over actions. The action-value function, which provides the expected return for taking an action $a$ in state $s$ under policy $\pi$, is given by:
\begin{equation}\label{eq2}
	Q^\pi \left( s, a \right) = \mathbb{E}^\pi \left[ \mathcal{R} \left( s^{(t)}, a^{(t)} \right) + \gamma V^\pi \left( s^{(t+1)} \right) \mid s^{(t)}, a^{(t)} \right]
\end{equation}

The optimal policy $\pi^*$ maximizes the state-value function for all states, leading to the optimal state-value function $V^*(s)$. Similarly, the optimal action-value function $Q^*(s, a)$ is obtained by following the optimal policy $\pi^*$. The Bellman equations for the optimal state-value and action-value functions are given by:
\begin{equation}\label{eq3}
	V^*(s) = \max_a \mathbb{E} \left[ \mathcal{R} \left( s^{(t)}, a^{(t)} \right) + \gamma V^* \left( s^{(t+1)} \right) \mid s^{(t)}, a^{(t)} \right]
\end{equation}
\begin{equation}\label{eq4}
	Q^*(s, a) = \mathbb{E} \left[ \mathcal{R} \left( s^{(t)}, a^{(t)} \right) + \gamma \max_{a^{(t+1)}} Q^* \left( s^{(t+1)}, a^{(t+1)} \right) \mid s^{(t)}, a^{(t)} \right]
\end{equation}

\subsection{Multi-Agent Markov Decision Process}
The standard multi-agent reinforcement learning (RL) model is formally defined as a 7-tuple $\langle\mathcal{N}, \mathcal{S}, \mathcal{O}, \mathcal{A}, P, R, \gamma\rangle$. Here, $\mathcal{N}$ represents the set of agents, with $|\mathcal{N}|$ indicating the number of agents. The state space is denoted by $\mathcal{S}$. Each agent $i$ in $\mathcal{N}$ can only observe a portion of the state $s \in \mathcal{S}$, represented by the observation $o_i$. The joint observation space is $\mathcal{O}=$ $\left\langle\mathcal{O}_1, \ldots, \mathcal{O}_{|\mathcal{N}|}\right\rangle$. Similarly, $\mathcal{A}_i$ represents the action space of agent $i$, and the joint action space is denoted as $\mathcal{A}=\left\langle\mathcal{A}_1, \ldots, \mathcal{A}_{|\mathcal{N}|}\right\rangle$. The transition function $P: \mathcal{S} \times \mathcal{A} \times \mathcal{S} \rightarrow[0,1]$ maps a state-action pair to a new state, reflecting the environment changes caused by agent actions. The reward function $\mathcal{R}_i: \mathcal{S} \times \mathcal{A} \rightarrow \mathbb{R}$ maps a state-action pair to a real value, representing the feedback received by agent $i$ due to the joint actions of all agents.

Formally, the objective of multi-agent reinforcement learning (MARL) in an MDP is to find a joint policy $\pi=\left\langle\pi_i, \ldots, \pi_{|\mathcal{N}|}\right\rangle$ such that each agent $i$ can maximize its expected discounted return. This is quantified by the state-value function:
\begin{equation}\label{eq5}
	V_i^\pi(s)=\mathbb{E}^\pi\left[\sum_{k=0}^{\infty} \gamma^k \mathcal{R}_i\left(s^{(t)}, \pi\left(s^{(t)}\right) \mid s_0=s^{(t+1)}\right)\right]
\end{equation}
where $\pi_i: O_i \rightarrow \Delta\left(A_i\right)$ is the individual policy of agent $i$, mapping the observation space to a probability distribution over the action space. The action-value function is given by:
\begin{equation}\label{eq6}
	Q_i^\pi\left(s, a\right)=\mathbb{E}^\pi\left[\mathcal{R}_i\left(s^{(t)}\right)+\gamma V_i^\pi\left(s^{(t+1)}\right) \mid s^{(t)}\right]
\end{equation}
A Markov game becomes a cooperative game when the learning goals of the agents are positively correlated, meaning $\forall i \neq j \in \mathcal{N}, V_i^\pi(s) \propto V_j^\pi(s)$. The objective of a cooperative Markov game is typically defined as the average expected discounted return of the team of agents.

Similarly, a networked Markov game (NMG) is a Markov game that includes the adjacency relationships among agents. It is formally defined as a 7-tuple $\langle\mathcal{G}(\mathcal{N}, \mathcal{E}), \mathcal{S}, \mathcal{O}, \mathcal{A}, P, R, \gamma\rangle$, where $(i, j) \in \mathcal{E}$ indicates that agents $i$ and $j$ are adjacent, meaning there is information sharing between them. Apart from $\mathcal{N}$ being replaced by the graph $\mathcal{G}(\mathcal{N}, \mathcal{E})$, the other elements in the tuple are defined in the same way as in a Markov game.

\begin{table}[H]
	\centering
	\caption{Definitions of the Traffic Network Components}
	\label{tab:network_definitions}
	\begin{tabular}{l|l}
		\hline Notation & Definition \\
		\hline $\mathcal{N}$ & Set of intersections \\
		\hline $\mathcal{N}_i$ & Set of intersection $i$'s neighbors \\
		\hline $\mathcal{E}$ & Edges between adjacent intersections \\
		\hline $L_i$ & Incoming lanes of $i$ \\
		\hline $\text{wait}_l$ & Number of waiting vehicles along incoming lane $l$ \\
		\hline $\text{phase}_i$ & Phase of intersection $i$ \\
		\hline $r_i$ & Individual reward of intersection $i$ \\
		\hline $o_i$ & Observation of agent $i$ \\
		\hline
	\end{tabular}
\end{table}

\section{Dependency Dynamics for MARL based TSC} \label{S4}
In this section, we start by proposing the concept of dependency dynamics for a networked Multi-Agent Markov Decision Process (MAMDP), followed by elaborating the mechanisms to utilize the dependency dynamics to mitigate the challenges of non-stationarity, scalability, and coordination complexity. Then, we show that in MARL-based TSC, the dynamic spill-back effect can be modeled as dependency dynamics and provide an analysis about the impact of the spill-back effect on the MARL learning process.

\subsection{Dependency Dynamics in a Networked MAMDP}

In dynamic environments like traffic networks, the dependencies among agents are not static but evolve over time based on factors such as traffic density, congestion levels, and agents' actions. To effectively model and address these changing dependencies, we introduce the concept of dependency dynamics.

\noindent\textbf{Definition 1} (Traffic Network). A traffic network can be defined as a graph $G(\mathcal{N}, \mathcal{E})$, where $i, j \in \mathcal{N}$ and $i j \in \mathcal{E}$ indicate that the two intersections $i, j \in \mathcal{N}$ are physically connected. Similarly, we denote $\mathcal{N}_i=\{j \in \mathcal{N} \mid(i, j) \in \mathcal{E}\}$ as the set of $i$ 's neighbors. Table \ref{tab:network_definitions} provides definitions of the various components of the traffic network.

\noindent\textbf{Definition 2} (Dependency Dynamics). Dependency dynamics refers to the temporal evolution of dependency among agents caused by time-varying states of agents. It captures how the influence of one agent's actions on another agent's state or reward changes over time due to environmental factors and agents' policies. Dependency dynamics are characterized by time-varying functions that quantify the degree of dependency between agents. These functions can be influenced by various factors, such as the volume of traffic flow between intersections, queue lengths, and occurrences of congestion phenomena such as spill-back.

\subsection{ Spill-back Dynamics in Traffic Networks} 
\begin{figure}[t]
	\centering
	\includegraphics[width=1\linewidth]{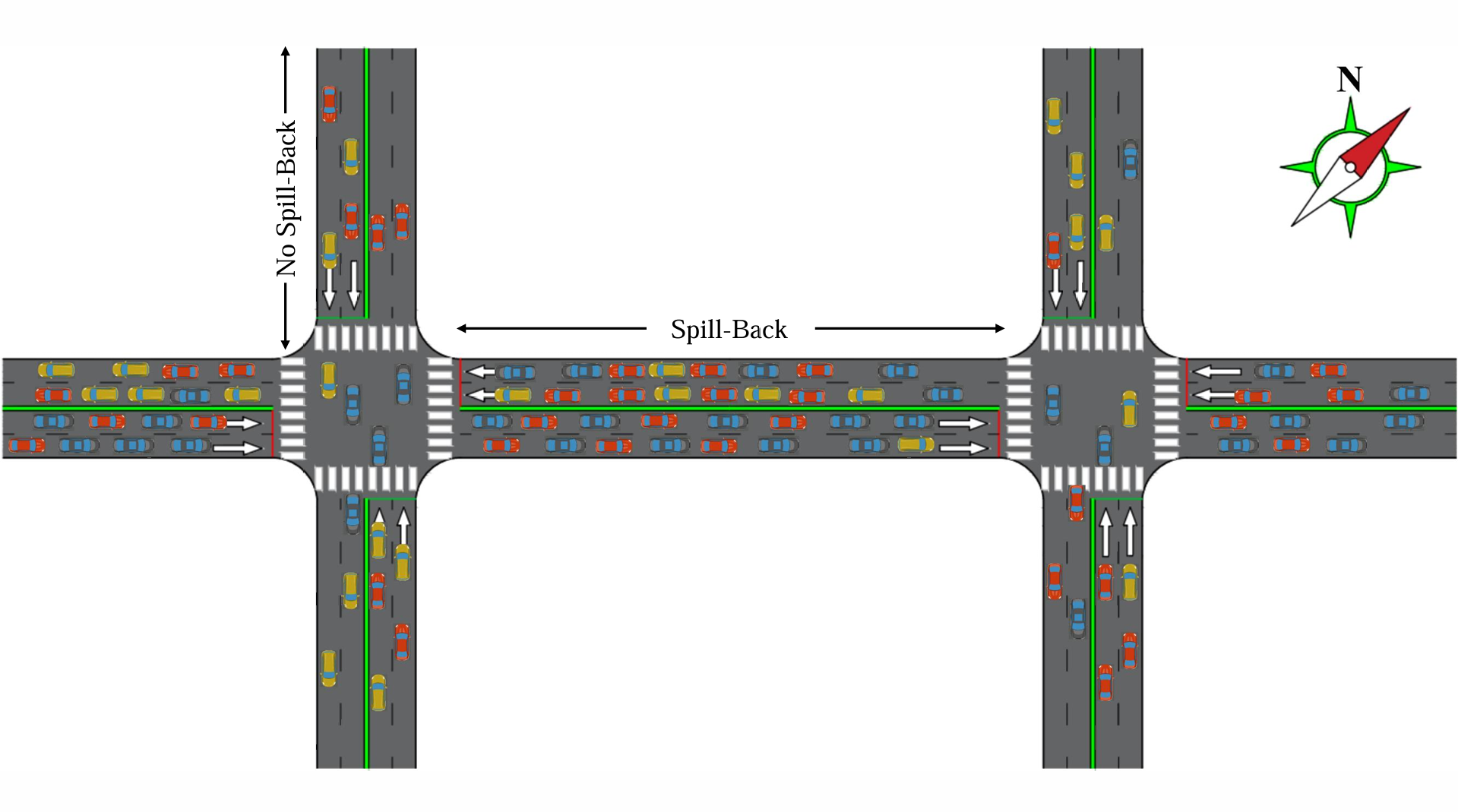}
	\caption{Spill-back process in ATSC.}
	\label{fig2}
\end{figure} 
Spill-back effect occurs when traffic congestion propagates to upstream road sections \cite{new5}. It happens when the queue at an intersection grows so long that it blocks the upstream intersection or roadway, preventing the free flow of traffic (Fig. 2). In Fig. 2, we can see that the congestion on the central lanes has caused a spill-back, blocking vehicles in the upstream sections, while the side lanes without spill-back allow vehicles to flow more freely.

Consider a multi-agent network $G(\mathcal{N}, \mathcal{E})$, where agents $i$ and $j$ are neighbors if there exists an edge $(i, j) \in \mathcal{E}$. The neighborhood of agent $i$ is denoted as $\mathcal{N}_i$, and the local region is defined as $\mathcal{V}_i=\mathcal{N}_i \cup\{i\}$. The distance $d(i, j)$ between two agents $i$ and $j$ is measured as the minimum number of edges connecting them. For instance, $d(i, i)=0$ and $d(i, j)=1$ for any $j \in \mathcal{N}_i$. In the Q-learning framework, each agent learns its policy $\pi$ and the corresponding value function $V_{w_i}$. The optimal action $a^*$ that maximizes the Q-value is determined by:
\begin{equation}\label{eq11}
	a^*_{i}=\operatorname{argmax} Q\left(\left(p_{L_i}^1, p_{L_i}^2, \ldots, p_{L_i}^{n_{L_i}}\right), a_{i}\right)
\end{equation}
where $ n_{L_i} $ denotes the number of cars on the incoming lane, and $p_{L_i}^c$ represents the position of the $c$-th car. Then, we can propose the following theorem:

\textit{Theorem 1:}
In the presence of spill-back, the congestion at one intersection can influence the state and actions of the agent at the neighboring intersection.

\textit{Proof for Theorem 1:}
 Formally, let $p_{L_i}^c$ represent the position of the $c$-th car on the incoming lane $L_i$ of intersection $i$, and similarly, $p_{L_j}^c$ for intersection $j$. The observations $o_i$ and $o_j$ of agents $A_i$ and $A_j$ are defined as:
\begin{equation}\label{eq12}
	o_i^{t}=\left(p_{L_i}^1, p_{L_i}^2, \ldots, p_{L_i}^{n_{L_i}}\right), \quad o_j^{t}=\left(p_{L_j}^1, p_{L_j}^2, \ldots, p_{L_j}^{n_{L_j}}\right)
\end{equation}
When spill-back occurs at intersection $i$, the effective state transition of the neighboring intersection $j$ is influenced. 
\begin{equation}\label{eq8}
	\mathcal{SP}_i= \begin{cases}1 & \text { if } \exists c \in\left\{1,2, \ldots, n_{L_i}\right\} \text { such that } p_{L_i}^c \leq P_{\text {th }} \\ 0 & \text { otherwise }\end{cases}
\end{equation}
Equivalently, it can be expressed as:
\begin{equation}\label{eq9}
	\mathcal{SP}_i=\max _{1 \leq c \leq n_{L_i}} \delta_{\left[p_{L_i}^c \leq P_{t h]}\right]}
\end{equation}

Let $\mathcal{SP}$ denotes the spill-back state of intersection $i$ and $j$. If $\mathcal{SP}_i=1$, the observation $o_j$ of agent $A_j$ changes due to the congestion propagated from intersection $i$ :
\begin{equation}\label{eq13}
	o_i^{(t+1)}=T\left(o_i^{(t)}, a_i^{(t)} \mid \mathcal{SP}_i\right)
\end{equation}
where $T$ is the transition function that now depends on the spill-back state $\mathcal{SP}_i$.
The joint state transition considering the influence of spill-back can be expressed as:
\begin{equation}
	\begin{gathered}
		\left(p_{L_i}^{1^{(t+1)}}, p_{L_i}^{2^{(t+1)}}, \ldots, p_{L_i}^{n_{L_i}^{(t+1)}}, p_{L_j}^{1^{(t+1)}}, p_{L_j}^{2^{(t+1)}}, \ldots, p_{L_j}^{n_{L_j}^{(t+1)}}\right)= 
		T\left(\left(p_{L_i}^{1^{(t)}}, p_{L_i}^{2^{(t)}}, \ldots, p_{L_i}^{n_{L_i}^{(t)}}, p_{L_j}^{1^{(t)}}, p_{L_j}^{2^{(t)}}, \ldots, p_{L_j}^{n_{L_j}^{(t)}}\right), (a_i^{(t)}, a_j^{(t)})\right)
	\end{gathered}
\end{equation}

Given that spill-back affects the next state of neighboring agents, the Q-value updates must consider these dependencies. The optimal $Q^*$ value for the agent $A_i$ at intersection $i$ from \eqref{eq4} becomes:
\begin{equation} \label{14}
	Q^*(o_i, a_i)=\mathcal{R}\left(o_i^{(t)}, a_i^{(t)}\right)+\gamma \max _{a_i^{(t+1)}} Q^*\left(o_i^{(t+1)}, a_i^{(t+1)}\right) 
\end{equation}
where $o_i^{(t+1)}$ is influenced by $o_j$ through spill-back state $\mathcal{SP}_i$ :
\begin{equation}\label{15}
	o_i^{(t+1)}= \begin{cases}T\left(o_i^{(t)}, a_i^{(t)} \mid \mathcal{SP}_i=1\right) \\T\left(o_i^{(t)}, a_i^{(t)} \mid \mathcal{SP}_i=0\right)\end{cases}
\end{equation}
Spill-back causes the same state to use the same action throughout the reinforcement learning process, and the state shifts to a different state, as shown in the following Eq. \ref{15}. By integrating Eq. \ref{14} and Eq. \ref{15}, it is evident that spill-back causes changes in the optimal Q-value, rendering the system non-deterministic.

\subsection{Impact of spill-back dynamics on MARL learning process}
The challenges of coordination complexity, communication overhead, and non-stationarity can be alleviated by decomposing MARL into multiple independent reinforcement learning (IRL) processes when no spill-back occurs across all intersections throughout the entire episode. Based on Definition 1 and Theorem 1, the following Corollary can be derived:

\textit{Corollary 1:} Equivalence of fully decentralized and fully centralized RL without spill-back process

\textit{Proof for Corollary 1:} Consider two intersections, $i$ and $j$, represented by agents $A_i$ and $A_j$, respectively, 
Let $\mathcal{N}$ be the set of all intersections, where each intersection $i \in \mathcal{N}$ is controlled by an agent $A_i$. Let $\mathcal{N}_i \subset \mathcal{N}$ denote the set of neighbors of intersection $i$. Let $s_i \in \mathcal{S}$ and $s_j \in \mathcal{S}$ be the states of intersections $i$ and $j$, respectively, and the joint state be $s=\left(s_i, s_j\right)$. The observations of agents $A_i$ and $A_j$ are $o_i \in \mathcal{O}_i$ and $o_j \in \mathcal{O}_j$, respectively, where $o_i$ and $o_j$ depend only on the local states $s_i$ and $s_j$, respectively, in a decentralized setting.

Let $a_i \in \mathcal{A}_i$ and $a_j \in \mathcal{A}_j$ be the actions taken by agents $A_i$ and $A_j$, respectively, with the joint action being $a=\left(a_i, a_j\right) \in \mathcal{A}$. The local rewards received by agents $A_i$ and $A_j$ are $r_i\left(s_i, a_i\right) \in \mathcal{R}$ and $r_j\left(s_j, a_j\right) \in \mathcal{R}$, respectively. The joint reward in a centralized setting is $r(s, a)=r_i\left(s_i, a_i\right)+r_j\left(s_j, a_j\right)$. The local transition probabilities for intersections $i$ and $j$ are $P\left(s_i^{(t+1)} \mid s_i, a_i\right)$ and $P\left(s_j^{(t+1)} \mid s_j, a_j\right)$, respectively, with the joint transition probability in a centralized setting being $P\left(s^{(t+1)} \mid s, a\right)=P\left(s_i^{(t+1)} \mid s_i, a_i\right) \cdot P\left(s_j^{(t+1)} \mid s_j, a_j\right)$, where $s^{(t+1)}=\left(s_i^{(t+1)}, s_j^{(t+1)}\right)$.

In the absence of spill-back, the state transitions and rewards at each intersection are independent. Consequently, the joint state $s=\left(s_i, s_j\right)$ transitions to $s^{(t+1)}=\left(s_i^{(t+1)}, s_j^{(t+1)}\right)$ as a result of the independent actions $a_i$ and $a_j$ executed by agents $A_i$ and $A_j$ respectively. Without spill-back, the traffic dynamics at intersection $i$ do not affect those at intersection $j$, and vice versa, implying that the state transitions and rewards are independent for the two intersections.
In a decentralized setting, the Q-value functions for agents $A_i$ and $A_j$ are defined as:
\begin{equation}
	\left\{\begin{array}{l}
		Q_i^{\mathrm{dec}}\left(s_i, a_i\right)=\mathbb{E}\left[r_i\left(s_i, a_i\right)+\gamma \sum_{s_i^{(t+1)}} P\left(s_i^{(t+1)} \mid s_i, a_i\right) V_i^{\mathrm{dec}}\left(s_i^{(t+1)}\right)\right] \\
		Q_j^{\mathrm{dec}}\left(s_j, a_j\right)=\mathbb{E}\left[r_j\left(s_j, a_j\right)+\gamma \sum_{s_j^{(t+1)}} P\left(s_j^{(t+1)} \mid s_j, a_j\right) V_j^{\mathrm{dec}}\left(s_j^{(t+1)}\right)\right]
	\end{array}\right.
\end{equation}
In a centralized setting, the joint Q-value function is defined as:
\begin{equation}\label{key}
	Q^{\text {cen }}(s, a)=\mathbb{E}\left[r(s, a)+\gamma \sum_{s^{(t+1)}} P\left(s^{(t+1)} \mid s, a\right) V^{\text {cen }}\left(s^{(t+1)}\right)\right]
\end{equation}
Substituting the joint reward and transition probabilities, we get:

\begin{equation}
	\begin{gathered}
		Q^{\operatorname{cen}}\left(\left(s_i, s_j\right),\left(a_i, a_j\right)\right)=\mathbb{E}\left[r_i\left(s_i, a_i\right)+r_j\left(s_j, a_j\right)+\right. 
		\left.\gamma \sum_{s_i^{(t+1)}} P\left(s_i^{(t+1)} \mid s_i, a_i\right) \sum_{s_j^{t+1}} P\left(s_j^{(t+1)} \mid s_j, a_j\right) V^{\operatorname{cen}}\left(s_i^{(t+1)}, s_j^{(t+1)}\right)\right]
	\end{gathered}
\end{equation}
Since the rewards and transitions are independent, the centralized value function can be decomposed as:
\begin{equation}\label{key}
	V^{\operatorname{cen}}\left(s_i^{(t+1)}, s_j^{(t+1)}\right)=V_i^{\mathrm{dec}}\left(s_i^{(t+1)}\right)+V_j^{\mathrm{dec}}\left(s_j^{(t+1)}\right)
\end{equation}
Therefore,
\begin{equation}\label{key}
	\begin{aligned}
		& Q^{\text {cen }}\left(\left(s_i, s_j\right),\left(a_i, a_j\right)\right)\\
		&=\mathbb{E}\left[r_i\left(s_i, a_i\right)+\gamma \sum_{s_i^{(t+1)}} P\left(s_i^{(t+1)} \mid s_i, a_i\right) V_i^{\text {dec }}\left(s_i^{(t+1)}\right)\right]+ 
		 \mathbb{E}\left[r_j\left(s_j, a_j\right)+\gamma \sum_{s_j^{(t+1)}} P\left(s_j^{(t+1)} \mid s_j, a_j\right) V_j^{\mathrm{dec}}\left(s_j^{(t+1)}\right)\right] \\
		& =Q_i^{\text {dec }}\left(s_i, a_i\right)+Q_j^{\text {dec }}\left(s_j, a_j\right)
	\end{aligned}
\end{equation}
Since the {Q}-values in the decentralized setting sum to the Q-value in the centralized setting, we conclude that, in the absence of spill-back, fully decentralized reinforcement learning is equivalent to fully centralized reinforcement learning.

In traffic theory, spill-back occurs when the queue length at an intersection extends upstream, interfering with traffic flow at preceding intersections. Without spill-back, traffic flow at each intersection is self-contained and does not impact neighboring intersections. This condition ensures that the learning process for traffic signal control at each intersection can be treated independently. Formally, the no spill-back condition is stated as:
\begin{equation}\label{key}
	\mathcal{SP}_i=0 \quad \text { and } \quad \mathcal{SP}_j=0
\end{equation}
where $\mathcal{SP}_i$ and $\mathcal{SP}_j$ are binary variables indicating the absence ( 0 ) of spill-back at intersections $i$ and $j$, respectively.

Under this condition, the state transitions at each intersection are independent:
\begin{equation}\label{key}
	T_i\left(s_i, a_i\right) \text { and } T_j\left(s_j, a_j\right)
\end{equation}
where $T_i$ and $T_j$ represent the state transition functions for intersections $i$ and $j$, respectively.

Therefore, without spill-back, the decentralized policies $\pi_i^{\mathrm{dec}}$ and $\pi_j^{\mathrm{dec}}$ learned by agents $A_i$ and $A_j$ are equivalent to the centralized policy $\pi^{\text {cen }}$ learned by a centralized agent controlling both intersections. This equivalence holds because the traffic dynamics at the intersections are independent, allowing for separate optimization of traffic signal control at each intersection without loss of generality.
\section{Deep Q-Network with Dynamic Parameter Update Strategy} \label{S5}
In this section, we will propose the DQN-DPUS, which can speed  up the convergence rate without sacrificing the optimal exploration by updating the weighting matrix and bias based on dependency dynamics.
\subsection{Steps of the DQN-DPUS algorithm}

\begin{algorithm}[h]
	\caption{DQN-DPUS Algorithm}
	\KwIn{ Q-network parameter: $\theta$, target Q-network parameters: $\theta^{-}$, learning rate: $\alpha$, discount factor: $\gamma$, buffer: $\mathcal{B}$, time horizon: $T$, max epochs: $M$, update frequency: $f$}
	\KwOut{Updated Q-network parameters: $\theta$}
	Initialize Q-network parameters $\theta$ and target Q-network parameters $\theta^{-} \leftarrow \theta$\;
	Initialize replay buffer $\mathcal{B}$\;
	\For{epoch = 1 to $M$}
	{
		\For{t = 1 to $T$}
		{
			Observe state $s^{(t)}$\;
			Select action $a^{(t)}$ using $\epsilon$-greedy policy from $Q(s^{(t)}, a^{(t)}; \theta)$\;
			Execute $a^{(t)}$, observe $r^{(t)}$, $s_{t+1}$\;
			Store transition $(s^{(t)}, a^{(t)}, r^{(t)}, s_{t+1})$ in buffer $\mathcal{B}$ with priority $p^{(t)}$\;
			\If{$t \% f == 0$}
			{
				\For{j = 1 to number of updates}
				{
				Compute TD-error $\delta = r + \gamma \max_{a^{(t+1)}} Q(s^{(t+1)}, a^{(t+1)}; \theta^{-}) - Q(s^{(t)}, a^{(t)}; \theta)$\;
				Update priorities $p_t \leftarrow |\delta|$\;
					\If{spill-back occurs}
					{
					$y = r + \gamma \max_{a^{(t+1)}} Q(s^{(t+1)}, a^{(t+1)}; \theta^{-})$\;
					$\theta \leftarrow \theta - \alpha \nabla_\theta (y - Q(s^{(t)}, a^{(t)}; \theta))^2$\;
							
					\ElseIf
					{
						$y = r + \gamma \max_{a^{(t+1)}} Q(s^{(t+1)}, a^{(t+1)}; \theta^{-})$\;
						$\theta_{\text{diag}} \leftarrow \theta_{\text{diag}} - \alpha \nabla_{\theta_{\text{diag}}} (y - Q(s^{(t)}, a^{(t)}; \theta_{\text{diag}}))^2$\;
					}

					}
				Periodically update $\theta^{-} \leftarrow \theta$\;
				}
					
			}
			
		}
	}
\end{algorithm}
\noindent\textbf{Step 1 Initialization:} Initialize the Q-network parameters $\theta$ and the target Q-network parameters $\theta^{-}$such that $\theta^{-} \leftarrow \theta$. Additionally, initialize the replay buffer $\mathcal{B}$ which will store the transitions observed by the agents during training. This buffer is crucial for experience replay, allowing the algorithm to learn from past experiences and improve stability and efficiency.\\
\textbf{Step 2 Training Loop:} Begin the main training loop that will iterate for a total of $M$ epochs. Each epoch represents a complete pass through the training process, during which the agent will interact with the environment multiple times. This iterative approach allows the agent to gradually learn and refine its policy over time by repeatedly experiencing different states and rewards.\\
\textbf{Step 3 Time Step Loop:} Within each epoch, iterate over a total of $T$ time steps. Each time step represents a single interaction with the environment, where the agent observes the current state, selects an action, and then observes the result of that action. This granular interaction is essential for learning the dynamics of the environment and optimizing the agent's actions. \\
\textbf{Step 4 Observe, Select, and Execute Action:} At each time step $t$, observe the current state $s_t$. Based on the observed state $s_t$, select an action $a_t$ using an $\varepsilon$-greedy policy derived from the Q-network $Q\left(s_t, a_t ; \theta\right)$. Execute the selected action $a_t$, and observe the immediate reward $r_t$ and the next state $s_{t+1}$. This feedback loop is critical for learning, as it provides the agent with real-time information about the consequences of its actions, enabling continuous improvement.\\
\textbf{Step 5 Store Transition:} If the current time step $t$ is a multiple of the predefined update frequency $f$, perform a series of updates on the Q-network. Sample a mini-batch of transitions $\left(s, a, r, s^{t+1}\right)$ from the replay buffer $\mathcal{B}$ based on their priorities. Sampling based on priority ensures that more significant transitions, which have a higher TD-error, are more likely to be selected for learning. This technique, known as prioritized experience replay, helps the agent learn more efficiently from important experiences, making the learning process more focused and effective.\\
\textbf{Step 6 Periodic Update Check and Mini-batch Sampling:} If the current time step $t$ is a multiple of the predefined update frequency $f$, perform a series of updates on the Q-network. Sample a mini-batch of transitions $\left(s, a, r, s^{t+1}\right)$ from the replay buffer $\mathcal{B}$ based on their priorities. Sampling based on priority ensures that more significant transitions, which have a higher TD-error, are more likely to be selected for learning.\\
\textbf{Step 7 Compute TD-error and Dynamic Update Parameters:} For each transition in the mini-batch, compute the temporal difference (TD) error $\delta$. The TD-error measures the difference between the predicted Q-value and the target Q-value. Update the priorities in the replay buffer based on the computed TD-errors. If a spill-back occurs, update all parameters $\theta$ of the Q-network considering the spill-back effect. Otherwise, update only the diagonal parameters $\theta_{\text {diag }}$ of the Q-network. This selective parameter update helps in efficiently learning the specific dynamics of spill-back situations.\\
\textbf{Step 8 Periodic Target Network Update and Output:} Periodically update the target Q-network parameters $\theta^{-} \leftarrow \theta$. This step helps in stabilizing the learning process by keeping the target network parameters fixed for a period of time, reducing the oscillations and divergence that can occur with rapidly changing targets. After completing the training process, return the updated Q-network parameters $\theta$. These parameters represent the learned policy of the agent, which can now be used to make decisions in the environment based on the learned Q -values. The final output is a Q-network that has been trained to optimize the agent's actions for maximum reward, considering both immediate and future consequences.

\begin{figure*}[t]
	\centering
	\includegraphics[width=1\linewidth]{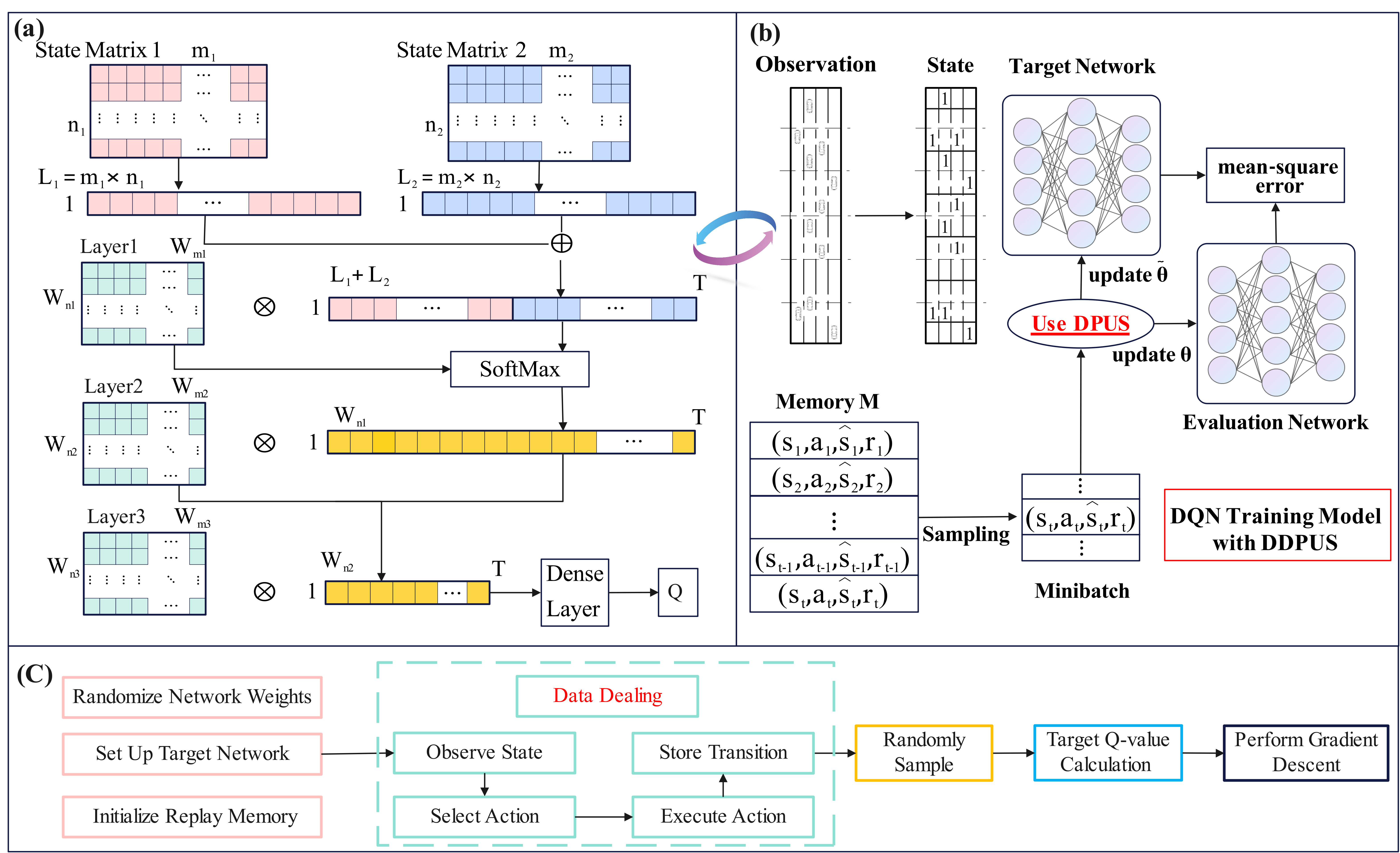}
	\caption{Overall architecture of the approach.}
	\label{fig4}
\end{figure*}

Fig. \ref{fig4} illustrates the training process of DQN-DPUS, where $W$ represents the network parameters during training, $m$ and $n$ denote the dimensions of the parameters, $T$ represents the transpose, and $L$ denotes the length.
\subsection{The effectiveness of the DQN-DPUS algorithm}
\textit{Definitions and Assumptions:}
Consider a system with two agents, $A_1$ and $A_2$, each controlling the traffic lights at two intersections. The state vector is $s=\left(s_1, s_2\right)$, and the action vector is $a=\left(a_1, a_2\right)$. The Q-value function is approximated by a neural network with a parameter matrix $W$, which is divided into four sub-matrices, and represented as follows:
\begin{equation}\label{eq23}
	W=\left(\begin{array}{ll}
		W_{11} & W_{12} \\
		W_{21} & W_{22}
	\end{array}\right)
\end{equation}
When there is no spill-back, the entire parameter matrix $W$ is updated. When there is spill-back, only the diagonal parameters $W_{11}$ and $W_{22}$ are updated.
We define a Lyapunov function $V(Q)$ to evaluate the change in Q-values:
\begin{equation}\label{key}
	V(Q)=\sum_{s, a}\left(Q(s, a ; W)-Q^*(s, a)\right)^2
\end{equation}
where $Q(s, a ; W)$ is the current Q-value and $Q^*(s, a)$ is the optimal Q-value. The goal is to minimize $V(Q)$ over whole episode.
When there is no spill-back, the update rule for the parameters is:
\begin{equation}\label{key}
	W \leftarrow W+\alpha_t \nabla_W L(W)
\end{equation}
where $\alpha^{(t)}$ is the learning rate and $L(W)$ is the loss function defined as:
\begin{equation}\label{key}
	\begin{split}
		L(W) &= \mathbb{E}_{\left(s^{(t)}, a^{(t)}, r^{(t)}, s^{t+1}\right) \sim D} \left[\left(r + \gamma \max_{a^{(t+1)}} Q\left(s^{(t+1)}, a^{(t+1)} ; W^{-}\right)  - Q(s^{(t)}, a^{(t)} ; W) \right)^2 \right]
	\end{split}
\end{equation}
The gradient of the loss function with respect to $W$ is given by:
\begin{equation}\label{key}
	\begin{split}
		\nabla_W L(W) &= \mathbb{E}_{\left(s^{(t)}, a^{(t)}, r^{(t)}, s^{(t+1)}\right) \sim D} \Bigg[2\left(r + \gamma \max_{a^{(t+1)}} Q\left(s^{(t+1)}, a^{(t+1)} ; W^{-}\right) - Q(s^{(t)}, a^{(t)} ; W)\right)\left(-\nabla_W Q(s^{(t)}, a^{(t)} ; W)\right)\Bigg]
	\end{split}
\end{equation}
When there is spill-back, the update rules for the diagonal parameters are:
\begin{equation}\label{key}
	\begin{gathered}
		W_{11} \leftarrow W_{11}+\alpha_t \frac{\partial L(W)}{\partial W_{11}} \\
		W_{22} \leftarrow W_{22}+\alpha_t \frac{\partial L(W)}{\partial W_{22}}
	\end{gathered}
\end{equation}
where the partial derivatives are:
\begin{equation}\label{key}
	\begin{aligned}
		\frac{\partial L(W)}{\partial W_{11}} &= \mathbb{E}_{\left(s^{(t)}, a^{(t)}, r^{(t)}, s^{(t+1)}\right) \sim D} \left[2\left(r + \gamma \max_{a^{(t+1)}} Q\left(s^{(t+1)}, a^{(t+1)} ; W^{-}\right)  - Q(s^{(t)}, a^{(t)} ; W)\right) \frac{\partial Q(s^{(t)}, a^{(t)} ; W)}{\partial W_{11}}\right] \\
		\frac{\partial L(W)}{\partial W_{22}} &= \mathbb{E}_{\left(s^{(t)}, a^{(t)}, r^{(t)}, s^{(t+1)}\right) \sim D} \left[2\left(r + \gamma \max_{a^{(t+1)}} Q\left(s^{(t+1)}, a^{(t+1)} ; W^{-}\right)-Q(s^{(t)}, a^{(t)} ; W)\right) \frac{\partial Q(s^{(t)}, a^{(t)} ; W)}{\partial W_{22}}\right]
	\end{aligned}
\end{equation}
Updating only the diagonal parameters $W_{11}$ and $W_{22}$ reduces the computational load for each update, thereby increasing the efficiency of the updates. By focusing on the parameters most relevant to the current state-action pairs, the updates are more precise, leading to faster convergence to the optimal values.

Using gradient descent theory, we can prove that under specific conditions, the diagonal parameter update strategy accelerates convergence. We analyze the expected change in the Lyapunov function as follows:
\begin{equation}\label{key}
	\begin{split}
		\mathbb{E}\left[V\left(Q^{(t+1)}\right) \mid Q^{(t)}\right] &\leq V\left(Q^{(t)}\right) - 2 \alpha_t \sum_{s, a} \left(Q(s^{(t)}, a^{(t)} ; W) - Q^*(s^{(t)}, a^{(t)})\right) 
		\quad \left(r^{(t+1)} + \gamma \max_{a^{(t+1)}} Q\left(s^{(t+1)}, a^{(t+1)} ; V\right)\right)
	\end{split}
\end{equation}
where $C$ is a constant representing the bound on the noise term in the update process. By updating only the diagonal parameters, unnecessary computations are avoided, leading to faster reduction in the Lyapunov function and thus accelerated convergence.

We can further decompose the expected change in the Lyapunov function:
\begin{equation}\label{key}
	\mathbb{E}\left[V\left(Q^{(t+1)}\right) \mid Q^{(t)}\right]=\mathbb{E}\left[\sum_{s, a}\left(Q^{(t+1)}(s^{(t+1)}, a^{(t+1)} ; W)-Q^*(s^{(t)}, a^{(t)})\right)^2 \mid Q^{(t)}\right]
\end{equation}
Using the update rule for no spill-back:
\begin{equation}\label{key}
	\begin{split}
		Q^{(t+1)}(s^{(t+1)}, a^{(t+1)} ; W) &= Q^{(t)}(s^{(t)}, a^{(t)} ; W) + \alpha_t \left( r + \gamma \max_{a^{(t+1)}} Q(s^{(t+1)}, a^{(t+1)} ; W^{-}) \right. 
		 \left. - Q^{(t)}(s^{(t)}, a^{(t)} ; W) \right)
	\end{split}
\end{equation}
We have:
\begin{equation}
	\begin{aligned}
		\mathbb{E}\left[V\left(Q^{(t+1)}\right) \mid Q^{(t)}\right]= & \mathbb{E}\Bigg[\sum_{s, a} \Big(Q^{(t)}(s, a ; W) \\
		&+ \alpha_t \Big(r + \gamma \max_{a^{(t+1)}} Q\left(s^{(t+1)}, a^{(t+1)} ; W^-\right)  - Q^{(t)}(s^{(t)}, a^{(t)} ; W) \Big) - Q^*(s^{(t)}, a^{(t)})\Big)^2 \mid Q^{(t)}\Bigg]
	\end{aligned}
\end{equation}
Simplifying, we get:
\begin{equation}
	\begin{aligned}
		\mathbb{E}\left[V\left(Q^{(t+1)}\right) \mid Q^{(t)}\right] \leq & V\left(Q^{(t)}\right)-2 \alpha_t \sum_{s, a}\left(Q^{(t)}(s^{(t)}, a^{(t)} ; W)-Q^*(s^{(t)}, a^{(t)})\right) \\
		& \times\left(r+\gamma \max _{a^{t+1}} Q\left(s^{(t+1)}, a^{(t+1)} ; W^{-}\right)-Q^{(t)}\left(s^{(t)}, a^{(t)} ; W\right)\right)\\
		&+\alpha_t^2 C
	\end{aligned}
\end{equation}

This shows that by updating only the diagonal matrices, the reduction in the Lyapunov function is more significant, leading to accelerated convergence.

\subsection{The convergence of the DQN-DPUS algorithm}
\textit{Definitions and Assumptions:} We define a MDP as a tuple $(\mathcal{S}, \mathcal{A}, P, r)$, where $\mathcal{S}$ is the (finite) state-space;
 $\mathcal{A}$ is the (finite) action-space;
 $P$ represents the transition probabilities;
 $r$ represents the reward function.

We denote elements of $\mathcal{X}$ as $x$ and $y$ and elements of $\mathcal{A}$ as $a$ and $b$. The reward is defined as a function $r: \mathcal{S} \times \mathcal{A} \times \mathcal{S} \rightarrow \mathbb{R}$, assigning a reward $r(x, a, y)$ every time a transition from $x$ to $y$ occurs due to action $a$. The reward $r$ is assumed to be a bounded, deterministic function.The value of a state $x$ is defined for a sequence of controls $\left\{A_t\right\}$ as:
\begin{equation}\label{key}
	J\left(x,\left\{A_t\right\}\right)=\mathbb{E}\left[\sum_{t=0}^{\infty} \gamma^t R\left(X_t, A_t\right) \mid X_0=x\right]
\end{equation}
The optimal value function is defined for each $x \in \mathcal{X}$ as:
\begin{equation}\label{key}
	V^*(x)=\max _{A_t} J\left(x,\left\{A_t\right\}\right)
\end{equation}
This verifies:
\begin{equation}\label{key}
	V^*(x)=\max _{a \in \mathcal{A}} \sum_{y \in \mathcal{X}} P_a(x, y)\left[r(x, a, y)+\gamma V^*(y)\right]
\end{equation}
From here, we define the optimal Q-function, $Q^*$ as:
\begin{equation}\label{key}
	Q^*(x, a)=\sum_{y \in \mathcal{X}} P_a(x, y)\left[r(x, a, y)+\gamma V^*(y)\right]
\end{equation}
In DQN-DPUS, the Q-value function is approximated by a neural network $Q(s, a ; \theta)$ with weights $\theta$. The target value is:
\begin{equation}\label{key}
	y^{(t)}=r^{(t)}+\gamma \max _{a^{(t+1)}} Q\left(s^{(t+1)}, a^{(t+1)} ; \theta^{-}\right)
\end{equation}
where $\theta^{-}$are the parameters of the target network, updated periodically to stabilize learning.
The loss function for the DQN algorithm is:
\begin{equation}\label{key}
	L(\theta)=\mathbb{E}_{\left(s^{(t)}, a^{(t)}, r^{(t)}, s^{(t+1)}\right) \sim \mathcal{D}}\left[\left(y_t-Q\left(s^{(t)}, a^{(t)} ; \theta\right)\right)^2\right]
\end{equation}
In the presence of spill-back, we update only the diagonal elements of the parameter matrix. Let $\theta_{\text {diag }}$ denote the diagonal elements of $\theta$. The update rule for the parameters is:
\begin{equation}\label{key}
	\theta_{\text {diag }, t+1}=\theta_{\text {diag }, t}+\alpha_t \nabla_{\theta_{\text {diag }}} L(\theta)
\end{equation}
The target network $Q\left(s, a ; \theta^{-}\right)$ is held fixed for a number of iterations, providing a stable target for learning. Periodic updates to $\theta^{-}$help in avoiding oscillations and divergence, a critical factor for ensuring convergence.
The use of experience replay ensures that the samples used for learning are independently and identically distributed., breaking the temporal correlations typically found in sequential data. This helps in approximating the theoretical conditions required for $\mathrm{Q}$-learning convergence.
The update rule in DQN-DPUS can be bounded similarly to Q-learning. Given the learning rate $\alpha_t$, the update step can be expressed as:
\begin{equation}\label{key}
	\Delta \theta_{\text {diag }, t}=\alpha_t \nabla_{\theta_{\text {diag }}} L(\theta)
\end{equation}
Ensuring $\sum_t \alpha_t=\infty$ and $\sum_t \alpha_t^2<\infty$ bounds the variance of the updates, allowing the parameters to converge over time.
To establish the convergence of DQN-DPUS, we present the following lemmas \cite{new6}:\\
\textbf{Lemma 1:} Given a finite MDP $(\mathcal{X}, \mathcal{A}, P, r)$, the DQN-DPUS, given by the update rule,
\begin{equation}\label{key}
	\theta_{\text {diag }, t+1}=\theta_{\text {diag }, t}+\alpha_t \nabla_{\theta_{\text {diag }}} L(\theta) \text {, }
\end{equation}
converges with probability 1 to a local optimum of the $\mathrm{Q}$-function as long as
\begin{equation}\label{key}
	\sum_t \alpha_t=\infty \quad \text { and } \quad \sum_t \alpha_t^2<\infty
\end{equation}
To establish \textbf{Lemma 1}, we need an auxiliary result from stochastic approximation, which we promptly present:\\
\textbf{Lemma 2:} The random process $\left\{\Delta_t\right\}$ taking values in $\mathbb{R}^n$ and defined as $\Delta_{t+1}(x)=\left(1-\alpha_t(x)\right) \Delta_t(x)+\alpha_t(x) F_t(x)$
converges to zero with probability 1 under the following assumptions:\\
- $0 \leq \alpha_t \leq 1, \sum_t \alpha_t=\infty$ and $\sum_t \alpha_t^2<\infty$;\\
- $\left\|\mathbb{E}\left[F_t(x) \mid \mathcal{F}_t\right]\right\| \leq \gamma\left\|\Delta_t\right\|$ with $\gamma<1$;\\
- $\operatorname{var}\left[F_t(x) \mid \mathcal{F}_t\right] \leq C\left(1+\|\right.$ Delta $\left._t \|^2\right)$, for $C>0$.\\
We start by rewriting the update rule as:
\begin{equation}\label{key}
	\theta_{\text {diag }, t+1}=\theta_{\text {diag }, t}+\alpha_t \nabla_{\theta_{\text {diag }}} L(\theta) \text {. }
\end{equation}
Subtracting $\theta_{\text {diag }}^*$ from both sides and letting $\Delta_{\text {diag }, t}=\theta_{\text {diag }, t}-\theta_{\text {diag }}^*$ we get:
\begin{equation}\label{key}
	\Delta_{\text {diag }, t+1}=\left(1-\alpha_t\right) \Delta_{\text {diag }, t}+\alpha_t \nabla_{\theta_{\text {diag }}} L(\theta) \text {. }
\end{equation}
If we write
\begin{equation}\label{key}
	F_t\left(\theta_{\text {diag }}\right)=\nabla_{\theta_{\text {diag }}} L(\theta)
\end{equation}
we have
\begin{equation}\label{key}
	\mathbb{E}\left[F_t\left(\theta_{\text {diag }}\right) \mid \mathcal{F}_t\right]=\nabla_{\theta_{\text {diag }}} L(\theta)-\nabla_{\theta_{\text {diag }}} L\left(\theta_{\text {diag }}^*\right)
\end{equation}
Using the fact that $\theta_{\text {diag }}^*$ is a local minimum of $L(\theta)$, we get
\begin{equation}\label{key}
	\mathbb{E}\left[F_t\left(\theta_{\text {diag }}\right) \mid \mathcal{F}_t\right]=\nabla_{\theta_{\text {diag }}} L(\theta)-\nabla_{\theta_{\text {diag }}} L\left(\theta_{\text {diag }}^*\right) \leq \gamma\left\|\Delta_{\text {diag }, t}\right\|
\end{equation}
Finally, the variance of $F_t\left(\theta_{\text {diag }}\right) \mid \mathcal{F}_t$ is
\begin{equation}\label{key}
	\begin{aligned}
		\operatorname{var}\left[F_t\left(\theta_{\text {diag }}\right) \mid \mathcal{F}_t\right] & =\mathbb{E}\left[\left(\nabla_{\theta_{\text {diag }}} L(\theta)-\nabla_{\theta_{\text {diag }}} L\left(\theta_{\text {diag }}^*\right)\right)^2 \mid \mathcal{F}_t\right] 
		 \leq C\left(1+\left\|\Delta_{\text {diag }, t}\right\|^2\right)
	\end{aligned}
\end{equation}
for some constant $C$. Then, by lemma 2, $\Delta_{\text {diag }, t}$ converges to zero with probability 1, i.e., $\theta_{\text {diag }, t}$ converges to $\theta_{\text {diag }}^*$ with probability 1 .
\section{ EXPERIMENTAL ENVIRONMENT} \label{S6}
In this section, we will describe the applied experimental environment for this research. We will firstly illustrate the road network model, followed by giving the definition of states, actions and reward.
\subsection{Scenario}
Our experimental scenario is a $1 \times 2$ road network, as shown in Figure 2. The distance between the two intersections is 300 meters. The traffic flow input for the experiment is illustrated in Figure 4. Figure 4 illustrates the throughput at two intersections over a period of 60 minutes, with the left column representing Intersection I and the right column representing Intersection II. Each subplot corresponds to different traffic input levels, with the traffic volume increasing from top to bottom.
\begin{figure}[t]
	\centering
	\begin{subfigure}{1\textwidth}
		\centering
		\includegraphics[width=\linewidth]{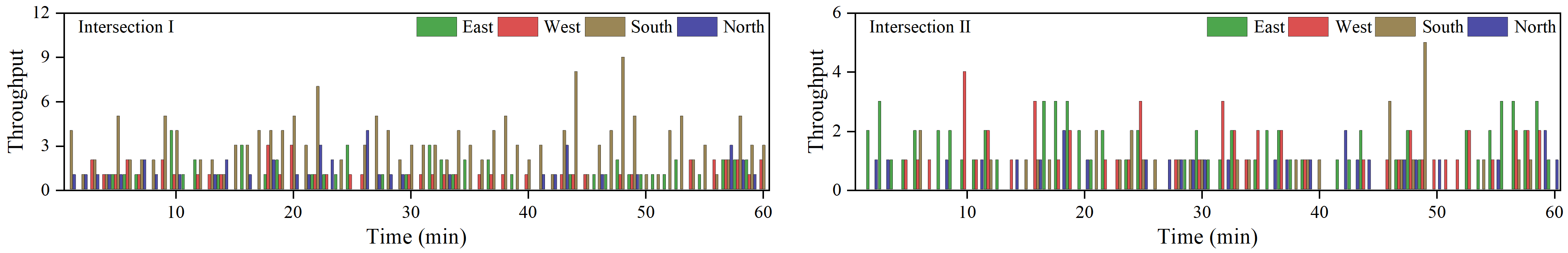}
		\label{fig:sub1}
	\end{subfigure}
	\vfill
	\begin{subfigure}{1\textwidth}
		\centering
		\includegraphics[width=\linewidth]{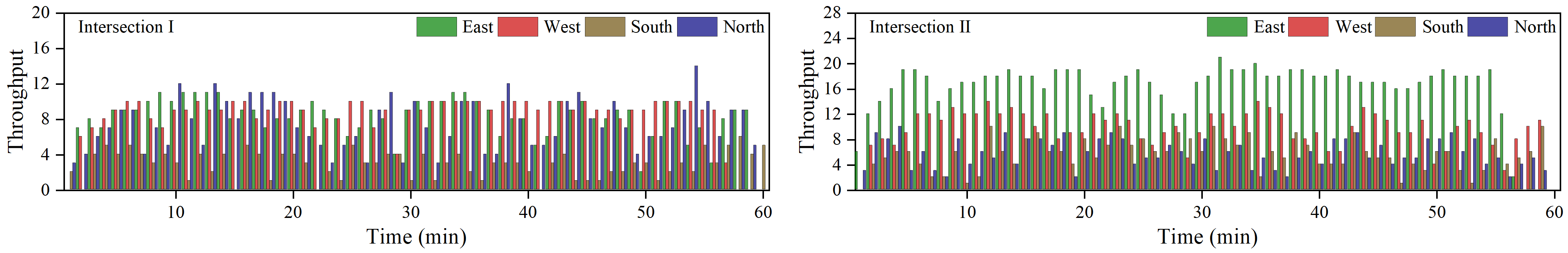}
		\label{fig:sub2}
	\end{subfigure}
	\vfill
	\begin{subfigure}{1\textwidth}
		\centering
		\includegraphics[width=\linewidth]{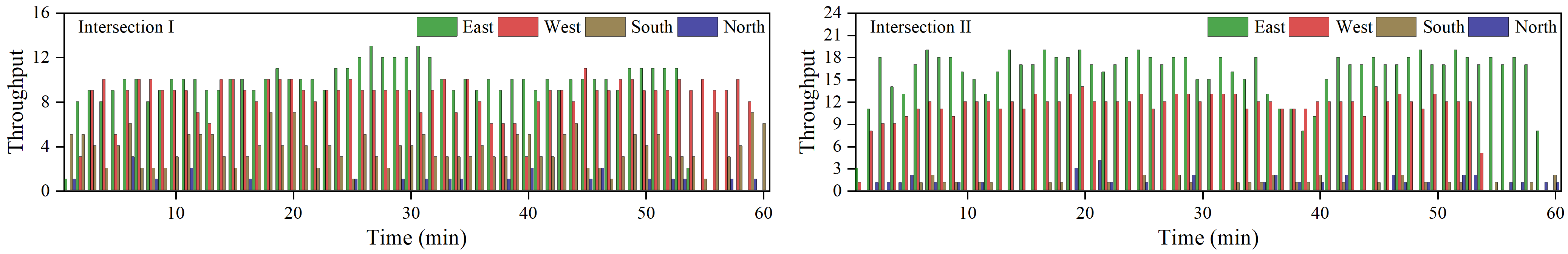}
		\label{fig:sub3}
	\end{subfigure}
	\vfill
	\begin{subfigure}{1\textwidth}
		\centering
		\includegraphics[width=\linewidth]{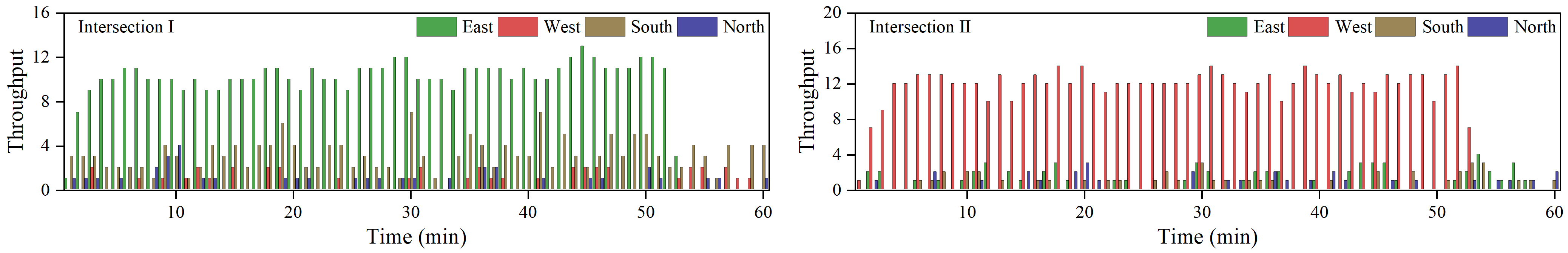}
		\label{fig:sub4}
	\end{subfigure}
	
	\caption{presents the traffic data stream we used (one hour), showing an increase in vehicle input from top to bottom. As the vehicle input increases, the degree of intersection spill-back also becomes more significant.}
	\label{fig3}
\end{figure}

\subsection{Definition of state, action and reward}
The state representation is defined as a matrix that captures the positions of vehicles within the road network. The road network is divided into a matrix, where cells containing a vehicle are marked as '1' and empty cells are marked as '0'.

The action space, denoted as $\mathcal{A}$, represents adjustments to the green light phase duration, specified by $a_{i}$. Possible actions include increasing, decreasing, or maintaining the current phase duration.

Consider that the objective of the ATSC problem is to enhance the traffic conditions within a specific region, with the count of vehicles waiting near intersections serving as a meaningful evaluation criterion. The cumulative rewards for all agents, observed following their actions at time $ t $, can be determined by computing the average number of waiting vehicles across all incoming lanes.
\begin{equation}\label{eq10}
	r_{i^{(t)}}=-\sum_{l \in L_i, i \in \mathcal{N}} \gamma^{t-1} \text { waiting time }[l]^{(t+1)}
\end{equation}
Since our goal is to minimize the number of waiting vehicles, we take a negative value in reward function.

\subsection{Baseline Methods}
We compare the performance of the DQN-DPUS with the following baseline methods, for which the algorithms are presented as follows:
s

(1) Multi-agent Advantage Actor-Critic (MA2C) [14]: We use A2C method separately at each intersection. Each agent uses a critic network to evaluate the policy of each actor and guide them to optimize their policies (fc is fully connected layer).
\begin{equation}
	\begin{aligned}
		\text {Actor, } \pi_{\textit{t}}^{\textit{i}}\left(\textit{s}_{\textit{t}}^{\textit{i}}\right) & =\operatorname{softmax}\left(f c\left(f c\left(\textit{s}_{\textit{t}}^{\textit{i}}\right)\right)\right) \\
		\text {Critic, } \textit{Q}_{\textit{t}}^{\textit{i}}\left(\textit{s}_{\textit{t}}^{\textit{i}}\right) & =\operatorname{relu}\left(f c\left(fc\left(\textit{s}_{\textit{t}}^{\textit{i}}\right)\right)\right)
	\end{aligned}
\end{equation}

(2) Fully Independent DQN Approach (In-DQN) [8]: We use the traditional DQN method in multi-agent reinforcement learning where each agent operates independently, without explicit coordination or communication with other agents.

(3) Cooperative Multi-agent Deep RL Approach (Co-MARL): Haddad et al., proposed a Co-MARL approach in 2022 \cite{new4}. The Co-MARL method applies a DQN, while transferring state, action and reward received from their neighbor agents to its own loss function during the learning process. Co-MARL uses the information transmitted by the neighbors as the state of learning to achieve the purpose of agent cooperation, the target value is:
\begin{equation}
	y_k^i=r_t^i+\gamma \cdot Q^i\left(s_{t+1}^i, \mathcal{S}_{t+1}^{\mathcal{G}_i}, a_{t+1}^i, \mathcal{A}_{t+1}^{\mathcal{G}_i} ; \theta_i\right)
\end{equation}


%
\begin{figure*}[t]
	\centering
	\includegraphics[width=1\linewidth]{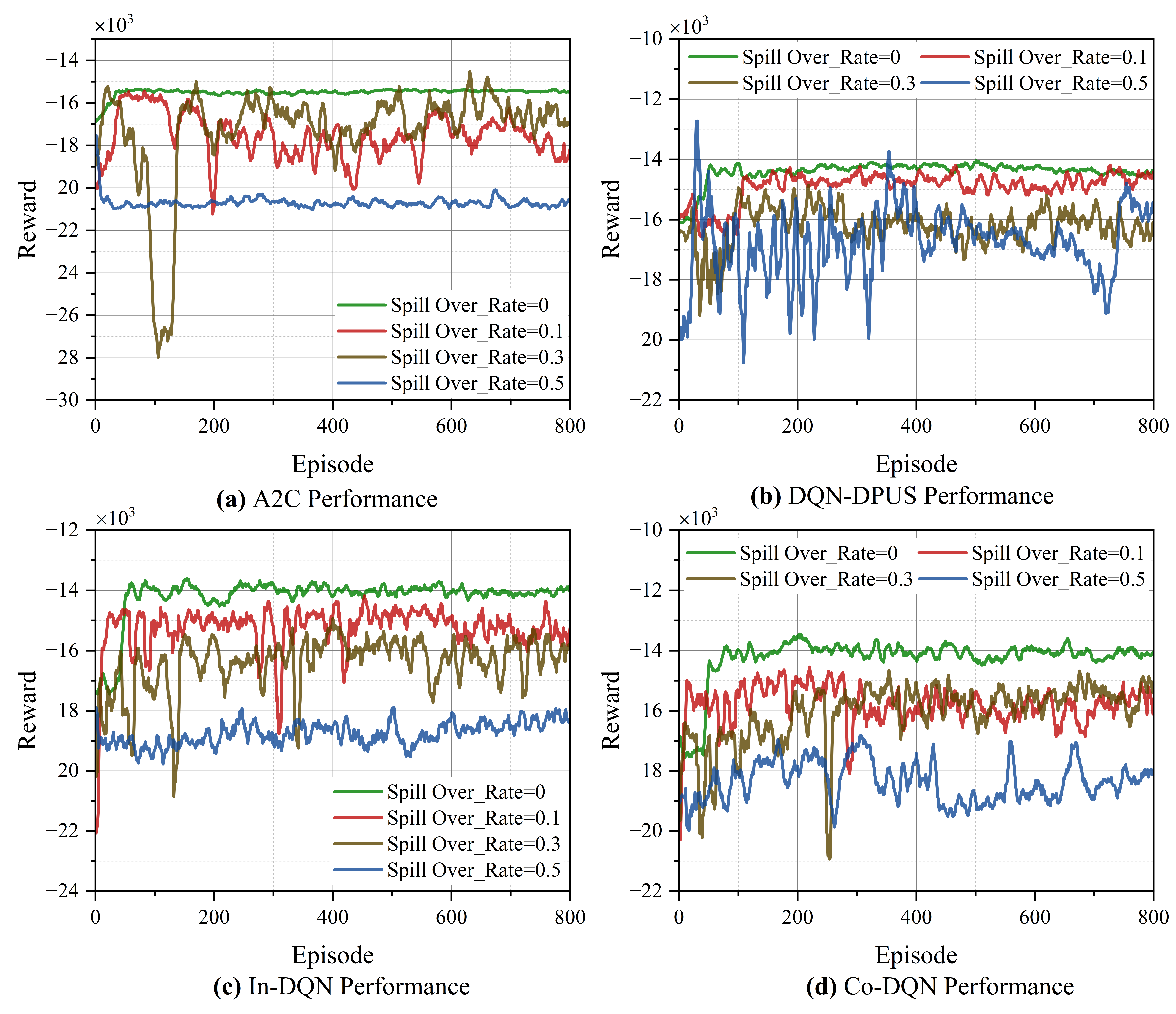}
	\caption{Performance of four algorithms under different spill-over rates.}
	\label{fig5}
\end{figure*}

\section{Result} \label{S7}
In this study, we compared the performance of four algorithms (A2C, DQN-DPUS, In-DQN, Co-DQN) under different spill-over rates (The frequency of vehicle spill-back occurrences in each episode.) (0, 0.1, 0.3, 0.5). The reward values were tracked over 800 episodes, with the primary metric being the average reward. The results show that the DQN-DPUS method outperforms the baseline methods across all spill-over rates, particularly under strong spill-over effects.

A detailed examination of the reward trends reveals several noteworthy observations. For instance, in the case of A2C at a spill-over rate of 0.5, there is a noticeable drop in performance around the 200th episode, followed by a period of recovery and subsequent fluctuations. This pattern indicates that A2C struggles to maintain stability in highly dynamic environments, reflected by the significant dips and spikes in rewards. Conversely, DQN-DPUS shows a more consistent performance trajectory across all spill- over rates, including 0.5, where the rewards stabilize after an initial learning phase and exhibit smaller fluctuations.

At the 600th episode, DQN-DPUS for a spill-over rate of 0.5 maintains a reward of approximately -18000, whereas A2C and In-DQN have rewards around -22000 and -21000, respectively. This clear margin underscores DQN-DPUS's ability to better manage strong spill-over effect. The reduced volatility in DQN-DPUS’s rewards at high spill-over rates suggests that this algorithm can adapt more effectively to changes in the environment, thereby providing more reliable performance.

Furthermore, when comparing the performance at a spill-over rate of 0.3, DQN-DPUS not only maintains higher average rewards but also exhibits lower variance compared to Co-DQN and In-DQN. For example, around the 400th episode, DQN-DPUS’s reward is approximately -17000, while Co-DQN fluctuates more widely around -18000. This consistency in DQN-DPUS’s performance can be attributed to its enhanced learning mechanisms, which allow it to handle the complexities introduced by spill-overs more robustly.

The robustness of DQN-DPUS is also evident when analyzing the initial learning phase (first 200 episodes). DQN-DPUS rapidly increases its reward values even in the presence of spill-over, indicating an efficient learning process. In contrast, In-DQN shows slower improvement and larger instability during this period, which can hinder their effectiveness in real-world applications where quick adaptation is crucial.

In summary, the analysis of specific points and trends within the reward trajectories demonstrates that DQN-DPUS not only achieves higher average rewards but also exhibits superior stability and robustness across varying spill-over rates. These qualities make DQN-DPUS a more reliable choice for environments characterized by high variability and complexity.

\section{Conclusion and Future Work} \label{S8}
The proposed DQN-DPUS algorithm effectively integrates prioritized experience replay and spill-back considerations to optimize traffic signal control in multi-agent environments. This algorithm dynamically adjusts the learning process based on real-time traffic conditions, allowing agents to make intelligent decisions that enhance traffic flow and reduce congestion. We demonstrated that spill-back effects lead to mutual influence between the intersections. Without spill-back, the fully centralized learning approach is equivalent to the fully independent learning algorithm. These experiments validate the effectiveness and convergence of the DQN-DPUS algorithm, highlighting its robustness in handling traffic scenarios. In future work, we plan to apply these approaches in vehicle-signal cooperative control. This will involve leveraging the developed methods to optimize the interaction between vehicles and traffic signals, aiming to improve overall traffic flow and reduce delays.

\clearpage

\bibliographystyle{elsarticle-num-names}

\bibliography{cas-refs}


\end{document}